\documentclass{article}

\usepackage[preprint]{neurips_2026}      % arXiv/preprint version

\usepackage[utf8]{inputenc}
\usepackage[T1]{fontenc}
\usepackage{hyperref}
\usepackage{url}
\usepackage{booktabs}
\usepackage{amsfonts}
\usepackage{amsmath}
\usepackage{nicefrac}
\usepackage{microtype}
\usepackage{xcolor}
\usepackage{graphicx}
\usepackage{float}
\usepackage{multirow}
\usepackage{wrapfig}
\usepackage{subcaption}
\usepackage{algorithm}
\usepackage{enumitem}
\usepackage{algpseudocode}
\usepackage{placeins}

\title{ Scaling Novel Graph Generation via Lightweight Structure-Guided Autoregressive Models}
\author{
Alessio Barboni \\
Boise State University \\
Boise, Idaho, USA \\
\texttt{alessiobarboni@u.boisestate.edu}
\And
Massimiliano Lupo Pasini \\
Oak Ridge National Laboratory \\
Oak Ridge, Tennessee, USA \\
\texttt{lupopasinim@ornl.gov}
\And
Bishal Lakha \\
Boise State University \\
Boise, Idaho, USA \\
\texttt{bishallakha@u.boisestate.edu}
\And
Edoardo Serra \\
Boise State University \\
Boise, Idaho, USA \\
\texttt{edoardoserra@boisestate.edu}
}

\begin{document}

\maketitle

%* insert abstract here * 

\begin{abstract}

Generating realistic and diverse graphs is a fundamental problem in machine learning, with applications spanning molecular and materials discovery, circuit design, cybersecurity tasks, and beyond. Despite rapid progress, existing graph generative models still face important limitations. Diffusion-based approaches often rely on costly operations over the full adjacency structure and long denoising chains, making generation computationally expensive and highly sequential in practice, while many autoregressive and hybrid models incur at least quadratic complexity in graph size. At the same time, across these model families, graph generators often remain strongly imitative, reproducing training patterns rather than exploring beyond the empirical training distribution. As a result, scalability is limited and genuine novelty remains difficult to achieve.

We address these issues with a lightweight autoregressive framework built around two key ideas. First, we introduce a structure-guided topological ordering that serializes each graph into a regular edge sequence, turning graph ordering from a nuisance into a computational advantage and enabling a near log-linear generation pipeline. Second, we propose a two-phase training strategy that reduces overfitting by combining exploration-oriented augmentation with iterative refinement. Together, these components encourage generation beyond the empirical training set while balancing novelty and plausibility.

Experiments across molecular benchmarks and a non-molecular large-graph setting show that the proposed approach maintains high validity and uniqueness while enabling controlled exploration beyond the training set. On QM9, it reaches 60.90\% VUN@10k against 54.99\% for the strongest baseline, and on MOSES 95.07\% VUN@25k against 85.38\%. Ablations further show that optimizing novelty without stability reaches 99.94\% novelty but reduces molecular stability to 2.77\%, whereas jointly selecting for novelty and stability retains 97.89\% novelty with 98.87\% molecular stability. We further show that the same serialization and training pipeline supports both LSTM and Mamba-style causal sequence backbones, and generates graphs on MalNet-Tiny, whose median training graph contains 2,272 nodes — a size regime in which dense diffusion baselines exhaust GPU memory during training.

\end{abstract}

\section{Introduction}

Many scientific and engineering systems are naturally represented as graphs.
Molecules can be modeled as atoms connected by chemical bonds, materials as
atomic structures, circuits as interconnected electronic components, and
software or cybersecurity artifacts as dependency or control-flow graphs. In
these domains, graph generation enables the synthesis of novel structured
objects that preserve the regularities observed in real data. Consequently,
graph generative models have become increasingly important in applications
ranging from molecular discovery to program analysis and cyber-defense.

Despite rapid progress, existing graph generative models still face important
limitations. Diffusion-based approaches often rely on costly operations over
the full adjacency structure and iterative denoising chains, making generation
computationally expensive and highly sequential in practice. Autoregressive
and hybrid approaches are often more lightweight, but their performance
strongly depends on the graph serialization strategy and many still incur at
least quadratic complexity in graph size. More fundamentally, across these
model families, graph generators frequently remain strongly imitative,
reproducing training examples or local structural patterns rather than exploring beyond the empirical training set while preserving plausible structure. As a result, achieving
both scalability and novelty remains challenging.

These issues become even more pronounced when graphs are represented in compact
discrete form. In this work, we encode graphs as ordered sparse edge streams
and model them at the bit level. Binary representations are attractive because
they are compact, memory efficient, and naturally scalable to sparse graphs.
However, they are also highly sensitive to node ordering: different
serializations of the same graph can produce drastically different binary
sequences, ranging from highly regular patterns to fragmented sequences that
are difficult for autoregressive models to learn.

To address these limitations, we propose a lightweight autoregressive graph
generation framework built around two key ideas. First, we introduce a
strict structure-guided serialization strategy that combines unsupervised structural node
representations derived from SIR-GN~\cite{joaristi2021sirgn} with a
breadth-first traversal of the graph. The structural representations rank
nodes according to their topological role, while the BFS traversal preserves
local connectivity during serialization. Together, these components transform
graph ordering from a nuisance into a computational advantage by producing
regular binary edge sequences that are substantially easier for autoregressive
models to learn. By generating sparse edge streams directly instead of dense
adjacency matrices, the framework achieves near-log-linear complexity for
sparse graphs, i.e., approximately $|E|\log|V|$, where $|V|$ and $|E|$ denote
the number of nodes and edges.

Second, we introduce a two-phase training strategy designed to balance
exploration and plausibility. In the first phase, the model is trained on both
original and structurally perturbed graphs, encouraging exploration beyond
simple memorization of the training data. In the second phase, we apply a
ReST-style refinement procedure~\cite{gulcehre2023rest}, where generated
graphs are filtered according to a Gaussian Mixture Model (GMM)  fitted in
graph embedding space generated by SIR-GN. The GMM models the distribution of training graphs in the SIR-GN embedding space, and generated graphs are retained according to a likelihood-based threshold before being used to further refine the generator. This refinement process encourages generation beyond the empirical training set while steering generated samples toward regions considered plausible in the learned embedding space.

The proposed framework supports both unsupervised and supervised graph generation settings. In the unsupervised setting, plausibility can be evaluated directly in embedding space without requiring domain-specific labels or handcrafted rules. When labels or task-specific evaluators are available, they can instead provide domain-dependent notions of validity and guide the refinement stage. More generally, embedding-space representations provide a natural way to extend notions of validity, uniqueness, and novelty beyond molecular graphs to domains in which explicit chemical validity rules are unavailable.

We evaluate the proposed framework on molecular and non-molecular benchmarks, including QM9, ZINC250K, MOSES %, ANI-1x, QM7x, Transition1x, 
and MalNet-Tiny. Across the molecular benchmarks, the proposed approach maintains high validity and uniqueness while enabling exploration beyond the training set; on QM9 it achieves 60.90\% VUN@10k compared with 54.99\% for the strongest baseline, and on MOSES 95.07\% VUN@25k compared with 85.38\%. Ablations further show that explicitly balancing novelty and molecular stability prevents the degradation that arises when novelty is optimized alone. On MalNet-Tiny, we additionally study the large-graph regime in which several recent graph generators encounter memory limitations. The results demonstrate that the proposed serialization and training framework can be used with lightweight causal sequence backbones, including recurrent LSTM decoders and Mamba-style state-space models.

%We evaluate the proposed framework on multiple molecular and non-molecular
%benchmarks, including QM9, ZINC250K, MOSES, ANI-1x, QM7x, Transition1x, and
% MalNet-Tiny. 
%We measure generation quality using standard metrics such as
% novelty, validity, and uniqueness. Novelty measures the fraction of generated
% graphs not present in the training set, validity measures the fraction of
% generated graphs satisfying domain-specific constraints, and uniqueness
% measures the proportion of non-duplicate generated samples. Experimental
% results show that the proposed approach substantially improves novelty while
% maintaining high validity and uniqueness, producing large numbers of novel
% graphs outside the training distribution with minimal duplication. 
% %The results demonstrate that controlled novelty and practical scalability can be achieved within lightweight autoregressive models such as LSTM or Linformer (Linear Transformers). 
% The results demonstrate that controlled novelty and practical scalability can be achieved within lightweight causal sequence backbones, including recurrent LSTM
% decoders and Mamba-style state-space models.

Our main contributions are summarized as follows:
\begin{itemize}[leftmargin=*, itemsep=0.00em, topsep=0.0em]

    \item We propose a compact bit-level edge serialization together with a structure-guided ordering that combines structural node ranking derived from SIR-GN with BFS traversal, producing regular binary sequences for autoregressive generation. This representation enables sparse graph generation with near-log-linear complexity, i.e.,
    $O(|E|\log|V|)$.

    \item We introduce a two-phase training framework that balances exploration
    and plausibility through graph perturbation and ReST-style embedding-space
    refinement. The framework supports both unsupervised and supervised
    selection criteria, and ablations demonstrate that the refinement stage can
    preserve molecular stability while encouraging generation beyond the
    empirical training set.

    \item We show empirically that serialization quality is decisive for bit-level autoregressive graph generation: fully random ordering causes a severe validity collapse, while plain BFS recovers much of the lost structure but is less stable than the full structure-guided traversal on harder datasets.

    \item We evaluate the framework across molecular and non-molecular graph
    benchmarks and with both LSTM and Mamba-style causal sequence backbones.
    The experiments demonstrate competitive joint validity--uniqueness--novelty
    performance and extend the evaluation to large sparse graphs where several
    recent graph generators encounter practical memory limitations.

\end{itemize}

\section{Related Work}

\paragraph{Autoregressive and Sequence-Based Graph Generation}

Autoregressive graph generators construct graphs through a sequence of local
decisions, such as adding nodes, predicting edges, or generating blocks of
structure. Early models such as DeepGMG and GraphRNN use recurrent networks to
generate graphs sequentially, with GraphRNN relying on BFS orderings to reduce
the space of possible node permutations \citep{li2018deepgmg,you2018graphrnn}. GRAN improves efficiency by generating
nodes and edges in blocks, while BiGG exploits graph sparsity to avoid dense
adjacency-matrix generation, reaching near-log-linear complexity through a recursive binary decomposition of the adjacency structure \citep{liao2019gran,dai2020bigg}. Our method reaches a comparable asymptotic regime by a different route: rather than decomposing the adjacency structure, we emit a flat bit-level stream of sorted edges under a fixed structural ordering, which keeps the generator a single causal sequence model and makes the backbone interchangeable.

A central limitation of autoregressive graph generation is that graphs do not
have a canonical linear order. As a result, the same graph may correspond to
many different sequences, and the chosen serialization strongly affects both
learning difficulty and generation quality. Recent work has revisited this
problem with more compact representations and larger sequence models. GEEL
uses a gap-encoded edge-list representation to reduce sequence and vocabulary
size, while G2PT uses graph tokenizations with a Transformer
decoder \cite{jang2024geel,chen2025g2pt}. AutoGraph similarly casts graph generation as sequence modeling, flattening graphs into token sequences that are generated autoregressively with a Transformer decoder \cite{chen2025autograph}. This provides a particularly relevant comparison to our approach: both reduce graph generation to causal sequence modeling, but our method instead uses structure-guided bit-level edge serialization and lightweight LSTM or state-space backbones. %Our method follows this graph-as-sequence view, but uses a compact bit-level edge serialization with lightweight causal sequence backbones.

\paragraph{Ordering and Serialization}

Because graph likelihoods are permutation-invariant but sequence likelihoods
are not, node ordering has become a central issue in autoregressive graph
generation. Earlier methods rely on fixed traversal or heuristic orderings,
such as BFS, DFS, degree-based orderings, or canonical ordering families. More
recent work treats ordering as a modeling problem. Order Matters models the
node sequence as a latent variable and uses variational inference to infer
likely generation orders \citep{chen2021ordermatters}. Latent Sort learns continuous embeddings whose
sorted order induces a graph permutation \citep{bu2023lettherebeorder}. LO-ARM introduces a trainable
state-dependent order policy that dynamically chooses the autoregressive order
and optimizes a variational lower bound \citep{wang2025loarm}. Our ordering instead uses SIR-GN structural node embeddings as an explicit
ranking signal, combining this structural ranking with a BFS constraint to
produce a connectivity-aware serialization \citep{joaristi2021sirgn}. This places our approach in the
broader line of structural node representation methods, such as struc2vec,
GraphWave, and SIR-GN, which represent nodes by structural role rather than
only proximity \citep{ribeiro2017struc2vec,donnat2018graphwave,joaristi2021sirgn}.

Other approaches reduce order sensitivity without learning a full order: orderless regularization penalizes recurrent hidden states that differ across valid flattenings of the same graph \cite{cohenkarlik2024orderless}, while MAG sidesteps next-node and next-edge ordering entirely by generating coarse-to-fine latent scales \cite{belkadi2025mag}. In contrast, we regularize the induced sequence through a fixed structural ranking rather than learning or marginalizing over generation orders.

% %Other methods attempt to reduce order sensitivity without learning a full
% order. Orderless regularization encourages recurrent hidden states to be
% similar across different valid graph flattenings, while MAG avoids explicit
% next-node or next-edge ordering by generating graphs through coarse-to-fine
% latent scales \cite{belkadi2025mag}. 
% % Our deterministic ordering is also connected to structural node representation
% % learning. Methods such as struc2vec, GraphWave, and SIR-GN embed nodes according
% % to structural role, providing signals that are complementary to traversal-based
% % locality. We use this type of structural signal to rank nodes before applying a
% % connectivity-aware BFS traversal.
% In contrast to these approaches, we regularize the induced sequence through a
% fixed structural ranking rather than by learning or marginalizing over orders.

\paragraph{Diffusion and Hybrid Models}

A second family of graph generators avoids dependence on a fixed serialization by modeling graphs more holistically. GraphVAE, EDP-GNN, GDSS, GraphGDP, and DiGress do so through latent-variable, score-based, or diffusion-based formulations \citep{simonovsky2018graphvae,niu2020edpgnn,jo2022gdss,huang2022graphgdp,vignac2023digress}. More recent diffusion and flow-based models refine this line: DisCo and Cometh formulate diffusion in continuous time over discrete graph states, DeFoG uses discrete flow matching to decouple training from sampling, and SparseDiff and Iterative Local Expansion target scalability through sparsity-preserving diffusion or localized generation \citep{qin2025sparsediff,xu2024disco,siraudin2024cometh,qin2025defog}. These formulations introduce different scaling costs: dense generators may reason over node pairs, giving quadratic dependence on graph size, while diffusion-based generators require repeated graph updates across sampling steps. We quantify these practical scaling limitations in Appendix~\ref{app:malnet_scalability}, where several recent graph generators encounter GPU-memory limits within the graph-size regime of MalNet-Tiny.

Hybrid models combine autoregressive and diffusion-based ideas. GraphARM adds nodes and incident edges through an autoregressive diffusion process, while PARD defines a structural partial order and generates graphs block by block with a diffusion decoder for each conditional block \citep{kong2023grapharm,zhao2024pard}. These approaches reduce dependence on a single generation order but require graph decompositions, block schedules, or repeated denoising. Our method instead keeps the generator purely autoregressive and lightweight.

\paragraph{Novelty-Oriented Training}

Most graph generative models are trained to match the empirical training
distribution through likelihood maximization, variational objectives, score
matching, flow matching, or denoising losses. These objectives are effective
for fidelity, but they do not necessarily encourage meaningful novelty. This tension is well known in molecular graph generation, where models are
commonly evaluated using validity, uniqueness, novelty, and distributional
metrics, which do not always move together.

Graph perturbations have also been widely used as augmentations in graph
representation learning, for example through edge dropping in DropEdge and
contrastive views in GraphCL \citep{rong2020dropedge,you2020graphcl}. In contrast, we use edge additions and removals
not to learn invariant representations, but to broaden the support of the
autoregressive generator during pretraining.

Our refinement stage is inspired by Reinforced Self-Training (ReST), which iteratively generates candidate samples and uses selected samples to improve the model \citep{gulcehre2023rest}, and is related to broader self-training and rejection-sampling fine-tuning procedures such as STaR and RAFT, where models are trained on filtered self-generated outputs \citep{zelikman2022star,dong2023raft}. Unlike these methods, which rely on a task reward or correctness label, our default acceptance criterion is label-free: Phase 2 retains generated graphs according to their likelihood under a GMM fitted in graph embedding space, which extends to domains where no domain-specific validity oracle exists. Where such an oracle is available, the same loop accepts domain-defined criteria instead, and we report both settings.

% Our refinement stage is inspired by Reinforced Self-Training (ReST), which
% iteratively generates candidate samples and uses selected samples to improve
% the model \citep{gulcehre2023rest}. This is also related to broader self-training and
% rejection-sampling fine-tuning procedures, such as STaR and RAFT, where models
% are trained on filtered self-generated outputs \citep{zelikman2022star,dong2023raft}. In our setting, the acceptance
% criterion is not a task reward, correctness label, or molecule-specific oracle:
% Phase~2 retains generated graphs that are plausible under a GMM fitted in graph
% embedding space.

Overall, our method differs from prior work in three ways: it uses structural
node representations to guide serialization, adopts a compact bit-level edge
representation, and explicitly separates exploration from exploitation through
perturbed pretraining and selective refinement. This combination is designed to
support scalable and novelty-oriented generation with a lightweight sequential
backbone.

\section{Methods}

\subsection{Problem Setup}

Let \( D = \{G_i\}_{i=1}^{N} \) be a dataset of attributed graphs. Each graph
\( G = (V, E) \) contains discrete node and edge attributes. Our goal is to
learn a generative model that produces novel graphs while remaining consistent
with the structural and attribute patterns observed in the training data.

We adopt an autoregressive formulation in which each graph is serialized into a
discrete sequence under a node ordering \( \pi(G) \). Because graphs in the
dataset may differ in size and connectivity, the resulting sequences generally
have different lengths. Denoting the serialized sequence by
\( s(G) = (x_1, \ldots, x_T) \), generation is modeled as
\[
p(s(G)) = \prod_{t=1}^{T} p(x_t \mid x_{<t}).
\]

In our setting, the node ordering used during serialization has a major impact
on learning, since it determines the sequence presented to the autoregressive
model and therefore the patterns the model must learn in order to predict the
next token.

\subsection{Structural Node Ordering}

We define a node ordering based on structural properties of the graph. This
procedure induces a partial ordering over the nodes, since structurally
equivalent nodes may receive the same rank. Our goal is not to compute a fully
canonical total ordering, but rather to reduce the ambiguity of graph
serialization by preferring node orders that are consistent with node
structure.

A purely structure-based ranking, however, does not by itself guarantee that
the serialization proceeds coherently within a connected component. Since our
generation process is intended to construct one connected component at a time,
we combine the structural ordering with a breadth-first search (BFS)
traversal. In this way, structural information determines node priority, while
BFS enforces a connectivity-aware exploration order.

More precisely, we first compute structural node representations using
SIR-GN, and use them to assign node ranks. During serialization, BFS
restricts the set of admissible next nodes to those on the current frontier
of the component being explored, while the structural ranking is used to
prioritize candidates within that set. Thus, the final ordering is not given
by the structural ranking alone, but by a structure-guided BFS traversal.

For graphs with multiple connected components, we apply the same procedure to
each component independently. After completing the BFS traversal of the current
component, a new component is initialized from the highest-ranked unvisited
node, and the process is repeated.

This ordering is particularly important in our setting because the downstream
generator operates on a binary serialization, whose learnability depends
strongly on the regularity of the induced sequence.

\subsection{Graph Serialization}

Given the structural node ordering described above, each graph is converted into
an ordered edge list, which is then flattened into a bit-level
representation for autoregressive modeling. After reindexing nodes according to the structural order, the edge list is
constructed so that edge pairs appear in strictly increasing lexicographic
order. This reduces arbitrariness in the serialization and yields a more
predictable sequence for the autoregressive model.

Each node appearing in an edge is represented through a fixed binary encoding.
The representation of an edge is then obtained by combining the binary
encodings of its two endpoints together with the corresponding edge attributes.
Concatenating these bit representations over the ordered edge list yields a
variable-length binary sequence $s(G)=(x_1,\dots,x_T)$, which serves as the target sequence for the autoregressive model.

The autoregressive decoder processes this flattened bit sequence one step at a time. In addition
to the previously generated bit, the model receives a positional encoding that
indicates which bit of the current edge representation is being predicted. This
provides the decoder with local context about its position inside the edge
encoding and helps stabilize sequence generation.

Although such binary encodings are often viewed as difficult to model due to their sensitivity to serialization choices, we find that the combination of structure-guided ordering and phased training makes this representation effective in practice.

\subsection{Autoregressive Sequence Backbones}
\label{sec:sequence-backbones}

We model the serialized bit sequence with a causal autoregressive decoder. In addition to next-bit prediction, the decoder predicts a stopping signal that determines when generation terminates. Training uses teacher forcing, while inference proceeds autoregressively until the stopping criterion is met.

% The ordered graph serialization described above is modeled using a causal
% autoregressive sequence decoder. Given the flattened bit sequence
% $s(G)=(x_1,\ldots,x_T)$, the model predicts each bit sequentially from the prefix
% observed so far,
% \[
%     p(s(G)) = \prod_{t=1}^{T} p(x_t \mid x_{<t}).
% \]
% At each step, the input consists of the previously generated bit together with a
% positional encoding indicating which position within the current edge
% representation is being predicted. This positional signal allows the decoder to
% distinguish between endpoint bits, edge-attribute bits, and stopping-related
% positions inside the serialized edge representation.

% The hidden representation at each step is used to predict the next bit in the
% sequence. In addition, the decoder predicts a stopping signal that determines
% when generation should terminate. During training, the model is optimized with
% teacher forcing, using the ground-truth previous bit at each step. At inference
% time, predictions are generated autoregressively until the stopping criterion is
% met.

\paragraph{LSTM backbone.}
Our primary implementation uses an LSTM decoder. This choice is intentional:
the LSTM provides a lightweight and transparent recurrent backbone, allowing us
to test whether the proposed structure-guided ordering, bit-level serialization,
and two-phase training procedure are sufficient for strong graph generation
without relying on a large or highly specialized architecture. Because the model
processes the serialized edge stream one bit at a time, the LSTM naturally
matches the causal structure of the generation problem.

\paragraph{Mamba backbone.}
We additionally evaluate a Mamba-based decoder for the long-sequence regime. Since each edge contributes $\Theta(\log |V|)$ bits, large sparse graphs such as MalNet-Tiny produce substantially longer sequences than molecular graphs. Although LSTM training is linear in sequence length, its recurrent transitions are inherently sequential and become a practical bottleneck at these lengths. Mamba \cite{gu2023mamba} instead uses a selective state-space model with a hardware-aware parallel-scan implementation, retaining linear sequence-length dependence while enabling more efficient training. We therefore use Mamba as an alternative backbone for long-sequence settings, while retaining the LSTM as the default on molecular benchmarks. %We therefore use Mamba for the long-sequence experiments and retain the LSTM as the default on molecular benchmarks.

% We also evaluate a Mamba-based decoder, motivated primarily by scalability rather
% than modeling capacity. Because each edge contributes $\Theta(\log |V|)$ bits to
% the serialized sequence, the total sequence length $T$ grows with both the number
% of edges and the graph size; large sparse graphs such as the MalNet-Tiny
% function-call graphs therefore produce bit streams orders of magnitude longer
% than those of small molecular graphs. For an LSTM decoder, teacher-forced
% training cost scales linearly in $T$ through a strictly sequential recurrence:
% each step depends on the previous hidden state, so the $T$ recurrent transitions
% cannot be parallelized along the sequence dimension. At the sequence lengths induced by MalNet-Tiny, this sequential dependence
% becomes the binding bottleneck: the per-epoch training time grows sharply, and
% the PyTorch/cuDNN LSTM backend does not support the longest tested
% MalNet-Tiny sequence configurations, as shown in Section~5.4. Mamba~\citep{gu2023mamba} replaces the recurrent
% transition with a selective state-space model whose recurrence admits a
% hardware-aware parallel-scan formulation, retaining linear dependence on $T$
% while substantially reducing the step-by-step training latency that limits the
% LSTM in this regime. We therefore use Mamba as the backbone for the
% long-sequence experiments, while retaining the LSTM as the default on the
% shorter molecular benchmarks, where it remains both competitive and more
% compact.

\paragraph{Autoregressive compatibility.}
We use Mamba causally, so predictions depend only on the available sequence prefix. When token representations are aggregated before prediction, we use prefix rather than full-sequence aggregation, preventing information leakage from future bits.

\subsection{Phase 1: Pretraining with Perturbed Graphs}

In the first phase, the autoregressive model is trained with teacher forcing on both the original training graphs and dynamically generated perturbations. At each epoch, we construct one perturbed graph per training example by randomly adding and removing edges, and serialize it using the same structure-guided ordering and bit-level encoding as the original graphs. The model is therefore trained on a mixture of observed and structurally modified graphs.

The purpose of this augmentation is exploratory rather than simply increasing the training-set size. By exposing the generator to nearby graph configurations beyond the empirical training set, Phase 1 broadens the sequence distribution encountered during training. Phase 2 subsequently provides the exploitative component by retaining generated samples according to an embedding-space notion of plausibility.

\subsection{Phase 2: ReST-style Refinement with GMM-based Filtering}

After pretraining, we refine the generator through iterative self-training inspired by ReST. At each iteration, the current model generates a set of candidate graphs, which are filtered before being used for subsequent teacher-forced training. During this phase, the original and perturbed Phase-1 graphs are no longer used.

To provide a domain-independent selection criterion, we fit a Gaussian Mixture Model (GMM) to embeddings of the original training graphs. Each generated graph is mapped into the same embedding space and scored under the GMM; samples below a quantile-based likelihood threshold are discarded, while the remaining samples are retained for continued training. This provides the exploitative component of the framework, selectively refining the exploratory generator toward regions considered plausible in the learned embedding space.

\section{Experimental Setup}

\subsection{Datasets}
\textbf{Datasets.} We evaluate on three molecular datasets—QM9, ZINC250K, MOSES—and on MalNet-Tiny, a non-molecular benchmark of Android malware function-call graphs. %For the quantum-chemical datasets, which may contain multiple conformations of the same molecule, we canonicalize molecules by SMILES and remove duplicate molecular graphs before training and evaluation so that novelty is measured at the graph level. 
For MalNet-Tiny, we consider one representative family as an initial test of non-molecular generation and scalability to substantially larger sparse graphs. %Dataset statistics and preprocessing details are provided in Appendix X.

\subsection{Evaluation Metrics}
\textbf{Metrics.} For molecular datasets, we report Validity, Uniqueness, and Novelty following standard practice. Validity is the fraction of chemically valid generated molecules, Uniqueness the fraction of non-duplicate samples, and Novelty the fraction not present in the training set. On QM9, we additionally report Atom and Molecule Stability. We also report VUN, the fraction of generated samples simultaneously satisfying validity, uniqueness, and novelty. Where applicable, we report standard distributional metrics such as SNN and scaffold similarity. Because some metrics are conditional on others (e.g., MOSES Novelty is evaluated over unique valid molecules), we interpret them jointly rather than in isolation. For MalNet-Tiny, we use representation- and classifier-based analogues of these metrics, defined in Section~\ref{sec:malnet_results}.

\subsection{Baselines}
\textbf{Baselines.} We compare against representative autoregressive, diffusion, flow-based, and hybrid graph generators: PARD, DiGress, DeFoG, AutoGraph, and SparseDiff. We additionally perform ablations of the proposed ordering and two-phase training procedure, including random and BFS orderings, training without Phase-1 perturbations, and training without Phase-2 refinement.

\FloatBarrier

\section{Results}
%Results show that we are better than all other methods. And that all components from the ordering, to the random graphs added, to the GMM were needed.
% Performance, Loss descent, \\

% Pretrain vs. no pretrain \\

% Score comparison w/ competitors, diff test metrics (le loro, la nostra) \\

% Raggiungiamo le performance di un link prediction AR brutale? (come baseline, su dati piccoli) \\

\subsection{Main Results}

\begin{table}[t]
\centering
\small

\caption{\textbf{QM9 molecular graph generation results.}
Higher is better. Our main results use the LSTM backbone unless otherwise stated. QM9 molecules are represented with explicit hydrogens. VUN@10k denotes the percentage of molecules, out of 10,000 generated samples, that are simultaneously valid, unique, and novel.}
\label{tab:qm9_results}
\setlength{\tabcolsep}{5.5pt}
\small
\begin{tabular}{lcccccc}
\toprule
Method
& Val. $\uparrow$
& Uniq. $\uparrow$
& Nov. $\uparrow$
& Atom. $\uparrow$
& Mol. $\uparrow$
& VUN@10k $\uparrow$ \\
\midrule
PARD        & 96.73 & 95.99 & 37.23 & 98.40 & 86.10 & 34.57 \\
DiGress     & 95.40 & \textbf{97.60} & 33.40 & 98.10 & 79.80 & 31.10 \\
DeFoG       & \textbf{98.23} & 96.28 & 36.00 & 98.99 & 91.00 & 34.04\\
AutoGraph   & 97.70 & 96.70 & 45.50 & 98.60 & \textbf{87.30} & 42.99 \\
SparseDiff  & 97.21 & 97.74 & 57.88 & 99.60 & 94.73 & 54.99 \\
\midrule
Ours - Labeled (VUNS) & 97.88 & 97.45 & \textbf{63.85} & \textbf{99.71} & \textbf{96.25} & \textbf{60.90} \\
Ours - Unlabeled & \textbf{98.23} & 92.10 & 59.25 & 98.75 &83.99 & 53.60 \\

% Ours (LSTM) - 1st epoch & \textbf{99.92} & 100.00 & 36.09 & 98.96 & 89.85 & 2.38 \\
% Ours (LSTM) - 99th epoch & 99.90 & 100.00 & \textbf{63.36} & \textbf{99.61} & \textbf{95.32} & 8.31 \\

%Ours -- ep.\ 1 & 94.97 & 94.28 & 29.85 & 98.07 & 85.67 & 1.63 & 2673 \\
%Ours -- ep.\ 49 & 96.19 & 93.31 & 79.79 & 97.31 & 54.84 & 5.77 & \textbf{7162} \\
%Ours -- ep.\ 99 & 84.42 & 92.50 & \textbf{88.74} & 95.43 & 39.59 & 6.53 & 6930 \\
\bottomrule
\end{tabular}
%\end{table}
\vspace{0.8em}

%\begin{table}[t]
%\centering
\captionof{table}{\textbf{ZINC250K molecular graph generation results.}
Higher is better. Our main results use the LSTM backbone unless otherwise stated.}
\label{tab:zinc_results}
%\setlength{\tabcolsep}{7pt}
%\small
\begin{tabular}{lccccc}
\toprule
Method
& Val. $\uparrow$
& Uniq. $\uparrow$
& Nov. $\uparrow$
& VUN@10k $\uparrow$ \\
\midrule
PARD & 98.68 & 99.95 & 99.69 & 98.33 \\
DiGress & 78.64 & 99.99 & \textbf{99.99} & 78.62  \\
DeFoG & \textbf{99.13} & \textbf{100.0} & 99.96 & \textbf{99.09} \\
AutoGraph & 87.68 & \textbf{100.0} & 95.50 & 83.73 \\
SparseDiff & 83.67 & 99.98 & 99.95 & 83.61 \\
\midrule
Ours - Labeled (VUN) & 97.19 & 99.28 & 99.94 & 96.43 \\
Ours - Unlabeled & 98.32 & 99.99 & 99.98 & 98.29 \\ %(LSTM) - ep.7
\bottomrule
\end{tabular}
%\end{table}
\vspace{0.8em}

%\begin{table*}[t]
\centering
\captionof{table}{\textbf{MOSES molecular generation results.}
Higher is better. `--' denotes metrics not reported or not
applicable. MOSES metrics evaluate validity, uniqueness, novelty, distributional
similarity, scaffold similarity, and filtering properties.}
\label{tab:moses_results}
\setlength{\tabcolsep}{4.2pt}
\small
\begin{tabular}{lccccccccccc}
\toprule
Method
& Val. $\uparrow$
& Uniq. $\uparrow$
& Nov. $\uparrow$
& Filters $\uparrow$
& SNN $\uparrow$
& Scaf. $\uparrow$
& VUN@25k $\uparrow$ \\
\midrule
PARD
& 86.80 & 100.00 & 78.20 & \textbf{99.0}
& \textbf{0.56} & 2.2 & 67.87 \\

DiGress
& 85.70 & 100.00 & 95.00 & 97.1
& 0.52 & 14.8 & 81.42 \\

DeFoG
& 92.80 & 99.90 & 92.10 & 98.9
& 0.55 & 14.4 & 85.38\\

AutoGraph
& 87.40 & 100.00 & 85.90 & 98.6
& 0.55 & -- & 75.08 \\

SparseDiff
& 84.70 & 100.00 & 95.10 & 97.0
& 0.52 & 15.5 & 80.55\\

\midrule
%Ours (LSTM) & \textbf{94.90} & 94.78 & \textbf{98.78} & 98.20 & 0.52 & \textbf{64.36} & 18.58 \\
%Ours (LSTM) & 90.64 & 97.44 & 93.49 & - & - & - & - \\
Ours - Labeled (VUN) & \textbf{97.81} & 98.47 & \textbf{98.71} & 87.64 & 0.41 & \textbf{21.8} & \textbf{95.07} \\
Ours - Unlabeled & 92.56 & 98.08 & 94.16 & 94.50 & \textbf{0.56} & 10.1 & 85.48 \\
\bottomrule
\end{tabular}
\end{table}

Tables 1–3 compare our method with recent graph generators on QM9, ZINC250K, and MOSES. On QM9, the labeled variant achieves 60.90\% VUN@10k, compared with 54.99\% for SparseDiff, while maintaining 97.88\% validity, 97.45\% uniqueness, and 96.25\% molecular stability. On MOSES, it reaches 95.07\% VUN@25k, compared with 85.38\% for DeFoG. On ZINC250K, both variants maintain high validity, uniqueness, and novelty, with the unlabeled variant reaching 98.29\% VUN@10k. The MOSES results also expose a fidelity–exploration tradeoff: the labeled variant achieves higher novelty and scaffold similarity, but lower Filters and SNN than several baselines, whereas the unlabeled variant reaches an SNN of 0.56.

\subsection{Ablation Summary}
\label{sec:ablation_summary}

We summarize the main ablation findings here and provide full curves and
discussion in Appendix~\ref{app:ablations}. Removing ReST after perturbed
pretraining leads to exploratory but poorly controlled generation, while
removing both perturbations and ReST produces a conservative model with
declining novelty. These results support the exploration--exploitation
interpretation of the two-phase procedure.

Ordering is also critical. Fully random ordering causes a severe validity
collapse, showing that bit-level generation is highly sensitive to
serialization. Plain BFS recovers much of the lost structure but is less stable
than the full structure-guided traversal on harder datasets. Appendix
\ref{app:ordering_analysis} further shows that the proposed structural ranking
yields more distinct node ranks than degree and PARD-style orderings.

Finally, removing perturbed graphs from Phase 1 makes refinement predominantly exploitative, further supporting the role of Phase-1 perturbations in encouraging exploration

% Finally, removing perturbed graphs from Phase~1 makes refinement predominantly
% exploitative.%, while replacing perturbations with BA graphs remains competitive.
% This suggests that Phase~1 benefits from structured exploratory pressure rather
% than one specific perturbation mechanism.

\FloatBarrier

\subsection{MalNet-Tiny target-family evaluation.}
\label{sec:malnet_results}
\begin{table}[!ht]
\centering
\caption{\textbf{Conditional generation on MalNet-Tiny.}
Validity is the fraction of generated graphs classified as the target
family by an external classifier trained only on real graphs.
Uniqueness and Novelty are computed in SIR-GN embedding space: %using two-decimal quantized embedding signatures: 
Uniqueness is the fraction of valid generations with distinct signatures, and Novelty is the
fraction of unique valid signatures not observed in the real
target-family training set. Higher is better.}
\label{tab:malnet_fidelity}
\small
\begin{tabular}{lcccc}
\toprule
Graphs / Method
& Validity $\uparrow$
& Uniqueness $\uparrow$
& Novelty $\uparrow$
& VUN@100 $\uparrow$ \\
\midrule
Real Family 0
& 88.0 & -- & -- & -- \\
Real Families 1--4
& 15.0 & -- & -- & -- \\
\midrule
AutoGraph
& \textbf{85.0}
& \textbf{77.6}
& 75.8
& 49.99 \\
\midrule
Ours (Phase 1)
& 74.0
& 68.9
& 56.9 
& 29.01 \\
Ours (Tempo 1)
& 64.0
& 68.8
& 59.1 
& 26.02 \\
Ours (Tempo 2)
& \underline{81.0}
& 67.9
& 81.8 
& 44.99 \\
Ours (Tempo 3)
& 70.0
& \underline{72.9}
& 90.2 
& 46.03 \\
Ours (Tempo 4)
& 77.0
& 72.7
& \textbf{96.4} 
& \textbf{53.96} \\
\bottomrule
\end{tabular}
\end{table}

% \begin{table}[t]
% \centering
% \caption{\textbf{Conditional generation on MalNet-Tiny.}
% We train a Family-0-vs-rest classifier on real training graphs and
% apply it to held-out real and generated graphs. The classifier achieves
% 86.5\% balanced accuracy on the real test set.}
% \label{tab:malnet_fidelity}
% \small
% \begin{tabular}{lc}
% \toprule
% Graphs & Classified as Family 0 (\%) $\uparrow$ \\
% \midrule
% Real Family 0 & 88.0 \\
% Real Families 1--4 & 15.0 \\
% \midrule
% AutoGraph & \textbf{85.0} \\
% Ours (Phase 1) & 74.0 \\
% Ours (Epoch 1) & 64.0 \\
% Ours (Epoch 2) & 81.0 \\
% Ours (Epoch 3) & 70.0 \\
% Ours (Epoch 4) & 77.0 \\
% \bottomrule
% \end{tabular}
% \end{table}

% We evaluate generated MalNet-Tiny graphs using a fixed external binary classifier trained only on real graphs to distinguish the target family from the remaining families. The classifier achieves 86.5\% balanced accuracy on held-out real graphs, with 88\% of real target-family graphs correctly classified as the target family, and is used only for evaluation. AutoGraph achieves a target-family classification rate of 85\%, while our model ranges from 64\% to 81\% across Phase-2 iterations, reaching its highest rate of 81\% at Epoch 2 (Table~4). The variation across iterations is non-monotonic, indicating that additional Phase-2 refinement does not consistently increase agreement with the target-family signal captured by the external evaluator.

For MalNet-Tiny, we define Validity as the fraction of generated graphs classified as the target family by the external classifier. To define representation-level Uniqueness and Novelty, we mean-pool the frozen SIR-GN node embeddings and quantize each embedding dimension, %to two decimal places 
yielding a discrete embedding signature. Uniqueness is the fraction of valid generations with distinct signatures, while Novelty is the fraction of unique valid signatures not observed among real target-family training graphs. Quantization avoids treating negligible floating-point differences as distinct representations.

% We evaluate generated MalNet-Tiny graphs using a fixed external binary classifier trained only on real graphs to distinguish the target family from the remaining families. The classifier achieves 86.86\% accuracy on held-out real graphs and is used only for evaluation, not for training or filtering. Across Phase-2 iterations, the fraction of generated graphs classified as the target family increases from 19\% to 78\% (Table 4), indicating progressively stronger agreement with the target-family signal captured by the external evaluator.

\textbf{Scalability.} MalNet-Tiny also exposes practical memory limitations of recent graph generators. On a GH200 with 95 GiB available GPU memory, DiGress exceeds memory at 1,750 nodes and SparseDiff at 2,999 nodes under the evaluated training configurations, while PARD inference fails at 5,000 nodes. Approximately 70\% of our MalNet-Tiny split exceeds 1,500 nodes, with a median training size of 2,272 nodes. Full configurations and stress-test results are reported in Appendix \ref{app:malnet_scalability}.

\FloatBarrier

\section{Limitations}

Several limitations remain. First, the method depends on the quality of the structure-guided ordering; if the ordering does not induce regular edge streams, bit-level autoregressive generation becomes substantially harder. Second, Phase 2 relies on a GMM fitted in graph-embedding space, so its filtering criterion is only as reliable as the embedding representation and may not capture all domain-specific notions of plausibility. Third, our MalNet-Tiny evaluation is an initial non-molecular feasibility study rather than a full benchmark: it considers one target family and evaluates target-family agreement using an external classifier trained on real graphs rather than a domain-specific oracle. A full multi-family evaluation remains future work. Finally, several runtime and backbone comparisons are single-run measurements and should therefore be interpreted as practical scaling evidence rather than definitive statistical benchmarks.

% Several limitations remain. First, the proposed method depends on the quality of
% the structure-guided ordering. If the ordering does not induce regular edge
% streams, bit-level autoregressive generation becomes substantially harder.
% Second, Phase~2 relies on a GMM fitted in graph-embedding space, so the filtering
% criterion is only as reliable as the embedding representation and may not capture
% all domain-specific notions of plausibility. Third, the current MalNet-Tiny experiment should be interpreted as an initial
% non-molecular feasibility study rather than a full benchmark. It focuses on one
% target family and evaluates structural fidelity through an external embedding
% classifier rather than through a domain oracle. A full multi-family MalNet-Tiny
% evaluation remains future work. Finally, several runtime and backbone comparisons
% are reported as single-run measurements rather than multi-seed averages, so they
% should be interpreted as practical scaling evidence rather than definitivec
% statistical benchmarks.

\section{Conclusion}

We introduced a lightweight autoregressive graph generator based on structure-guided bit-level edge serialization and two-phase refinement. Combining SIR-GN structural rankings with BFS traversal produces regular sparse edge sequences modeled at $O(|E|\log|V|)$ complexity, while perturbation-based pretraining and ReST-style GMM filtering balance exploration and plausibility. Across molecular benchmarks, the framework achieves competitive joint validity, uniqueness, and novelty, including 60.90\% VUN@10k on QM9 and 95.07\% VUN@25k on MOSES. Experiments on MalNet-Tiny further demonstrate generation in a substantially larger sparse-graph regime where several recent graph generators encounter practical memory limitations. These results highlight graph serialization as an important design axis for scalable, novelty-oriented autoregressive generation.

\section*{Acknowledgment}
This research was sponsored in part by the Artificial Intelligence Initiative as part of the Laboratory Directed Research and Development Program of Oak Ridge National Laboratory, managed by UT-Battelle, LLC, for the U.S. Department of Energy under contract DE-AC05-00OR22725. This work was also supported in part by the Supermicro NVIDIA Grace Enablement Evaluation Program by providing a Supermicro ARS-111GL-NHR server equipped with an NVIDIA GH200 Grace Hopper Superchip, which supported the evaluation and research activities reported in this paper.

%* insert conclusions here * 
%\end{document}  % This is where a 'short' article might terminate

%ACKNOWLEDGEMENTS are optional
%\section{Acknowledgements}

%* insert ACKs here *

\bibliographystyle{plainnat}
\bibliography{graph_generation_references}
% ACM needs 'a single self-contained file'!
% SIGKDD: balancing columns messes up the footers: Sunita Sarawagi, Jan 2000.
% \balancecolumns

%\input{checklist.tex}
\FloatBarrier
\appendix

\section{Novelty–Stability Selection Ablation}
We further examine the trade-off induced by the selection criterion used during
refinement. Selecting samples based on novelty alone encourages aggressive
exploration, but can substantially reduce molecular stability. Incorporating
stability into the selection criterion preserves most of the novelty gain while
producing substantially more stable molecules. Tables~\ref{tab:qm9_ablation} and
\ref{tab:qm9_ablation_dynamics} report the corresponding results.

\begin{table}[t]
\centering
\caption{Ablation of novelty ($N$) and stability ($S$) selection on QM9 after
20 Phase-2 epochs. All methods include validity ($V$) and uniqueness ($U$)
selection.}
\label{tab:qm9_ablation}
\resizebox{\columnwidth}{!}{
\begin{tabular}{lcc|ccccc}
\toprule
\textbf{Method} 
& $\mathbf{N}$ 
& $\mathbf{S}$
& \textbf{Val.} $\uparrow$
& \textbf{Uniq.} $\uparrow$
& \textbf{Nov.} $\uparrow$
& \textbf{Atom Stab.} $\uparrow$
& \textbf{Mol. Stab.} $\uparrow$ \\
\midrule
VU   & -- & -- & 99.21 & \textbf{92.95} & 98.67 & 80.57 & 13.88 \\
VUN  & \checkmark & -- & \textbf{99.74} & 92.92 & \textbf{99.94} & 71.48 & 2.77 \\
VUS  & -- & \checkmark & 99.67 & 88.20 & 87.02 & \textbf{99.96} & \textbf{99.56} \\
VUNS & \checkmark & \checkmark & 99.20 & 88.55 & 97.89 & 99.89 & 98.87  \\
\bottomrule
\end{tabular}
}
\end{table}

\begin{table}[t]
\centering
\caption{Evolution of novelty and molecular stability during Phase-2 training.
Novelty selection rapidly increases exploration, while stability selection
prevents the associated collapse in molecular stability.}
\label{tab:qm9_ablation_dynamics}
\resizebox{\columnwidth}{!}{
\begin{tabular}{lrrrrrrrr}
\toprule
& \multicolumn{2}{c}{\textbf{Epoch 1}}
& \multicolumn{2}{c}{\textbf{Epoch 3}}
& \multicolumn{2}{c}{\textbf{Epoch 10}}
& \multicolumn{2}{c}{\textbf{Epoch 20}} \\
\cmidrule(lr){2-3}
\cmidrule(lr){4-5}
\cmidrule(lr){6-7}
\cmidrule(lr){8-9}
\textbf{Method}
& \textbf{Nov.} & \textbf{Mol. Stab.}
& \textbf{Nov.} & \textbf{Mol. Stab.}
& \textbf{Nov.} & \textbf{Mol. Stab.}
& \textbf{Nov.} & \textbf{Mol. Stab.} \\
\midrule
VU
& 28.90 & 83.21
& 38.92 & 80.02
& 73.98 & 55.61
& 98.67 & 13.88 \\

VUN
& 43.30 & 78.85
& 96.95 & 7.84
& 99.65 & 3.78
& 99.94 & 2.77 \\

VUS
& 25.98 & 92.77
& 32.66 & 97.12
& 59.11 & 98.56
& 87.02 & 99.56 \\

VUNS
& 39.10 & 92.62
& 63.85 & 96.25
& 92.11 & 98.26
& 97.89 & 98.87 \\
\bottomrule
\end{tabular}
}
\end{table}

\section{Complexity Analysis}
\label{app:complexity}

We analyze the per-graph time complexity of the proposed pipeline. Let
$n=|V|$ and $m=|E|$. We focus on sparse graphs with $m=O(n)$, as in the
molecular and structural benchmarks considered in this work, and treat the
LSTM hidden size, SIR-GN iteration count, edge-attribute width, and graph
embedding dimension as constants.

\paragraph{Ordering.}
Computing structural node representations with SIR-GN requires linear work in
the graph size per iteration, hence $O(n+m)$ for a constant number of
iterations. Ranking the resulting node representations costs $O(n\log n)$.
The structure-guided BFS traversal uses a FIFO queue and visits each node and
edge a constant number of times; structural ranking is used only to order
neighbors before they are enqueued. Sorting neighbor lists contributes
$\sum_v O(\deg(v)\log \deg(v)) \leq O(m\log n)$. Thus, ordering costs
$O(n\log n + m\log n)$, which is $O(n\log n)$ for sparse graphs.

\paragraph{Serialization.}
Each node index requires $b=\lceil \log_2 n\rceil$ bits, so each edge
contributes $O(\log n)$ bits to the sequence. Lexicographic edge sorting costs
$O(m\log m)$, and writing the bit sequence costs $O(m\log n)$. The resulting
sequence length is $T=\Theta(m\log n)$, which is $O(n\log n)$ for sparse
graphs.

\paragraph{Sequence model.}
A forward or sampling pass through the LSTM costs $O(Td^2)$ for hidden size
$d$. Treating $d$ as fixed and using $T=O(n\log n)$ on sparse graphs, training
and autoregressive generation are near-log-linear per graph. This avoids
constructing or attending over the full $n\times n$ adjacency matrix, a common
bottleneck in dense graph generators. For the Mamba variant, the causal state-space backbone has linear dependence on
sequence length for fixed hidden dimension, so the sequence-length dependence
remains proportional to \(T = O(m \log n)\).

\paragraph{Phase 1.}
At each epoch, the model trains on the original graphs and one perturbed graph
per original graph. The perturbation step removes
$O(p_{\mathrm{drop}}n)$ edges and adds $O(p_{\mathrm{add}}n)$ candidate edges
for constant perturbation rates. Since added edges are sampled by rejection
rather than by constructing the full non-edge set, this step is linear in
expectation for sparse graphs. The dominant per-epoch cost is therefore
$O(|\mathcal{D}|\, n_{\max}\log n_{\max})$.

\paragraph{Phase 2.}
Each ReST iteration samples $M$ candidate graphs, scores them under the fitted
GMM, and trains on the accepted subset $\mathcal{C}_k$. Generating one candidate
costs $O(n_{\max}\log n_{\max})$. Computing a linear-time graph embedding adds
$O(n_{\max})$, and GMM scoring with $C$ components costs $O(C)$ for fixed
embedding dimension. Since $|\mathcal{C}_k|\leq M$, the dominant term per
iteration is $O(M\,n_{\max}\log n_{\max})$.

\paragraph{Fitting the GMM.}
The GMM is fit once before Phase 2 on training graph embeddings. Standard EM
with $C$ components and $|\mathcal{D}|$ samples costs
$O(I\,C\,|\mathcal{D}|)$ for $I$ iterations when the embedding dimension is
fixed. This is a one-time preprocessing cost.

\begin{table}[!htbp]
\centering
\small
\caption{Per-graph generation complexity of representative graph generative
models on sparse graphs ($m=O(n)$). $T_{\mathrm{steps}}$ denotes the number of
denoising, diffusion, or flow-matching sampling steps. Our general cost is
$O(m\log n)$; the table reports the sparse case.}
\label{tab:complexity}
\begin{tabular}{lll}
\toprule
Family & Representative method & Generation cost \\
\midrule
Dense diffusion / flow & DiGress, GDSS, DeFoG & $O(T_{\mathrm{steps}}\cdot n^2)$ \\
Hybrid AR--diffusion & PARD, GraphARM & $O(T_{\mathrm{steps}}\cdot n^2)$ \\
Dense AR & GraphRNN, GRAN & $O(n^2)$ \\
Sparse AR & BiGG & $O((n+m)\log n)$ \\
Sparse AR bit stream & \textbf{Ours} & $O(n\log n)$ \\
\bottomrule
\end{tabular}
\end{table}

\paragraph{Comparison with prior work.}
Table~\ref{tab:complexity} summarizes representative generation costs. Dense
adjacency-based autoregressive and diffusion models scale at least
quadratically in graph size, with diffusion or flow-based methods also
requiring multiple sampling steps. Sparse autoregressive models avoid this by
exploiting edge sparsity. Our method follows this sparse regime through a
bit-level edge stream, giving $O(m\log n)$ generation cost in general and
$O(n\log n)$ on sparse graphs.

\section{Competitor Scalability on MalNet-Tiny}
\label{app:malnet_scalability}
\textbf{DiGress.} We evaluated the original DiGress Graph Transformer with batch size 1 on an NVIDIA GH200 with 95 GiB of available GPU memory. A forward/backward pass required 83.64 GiB at 1,500 nodes and exceeded available memory at 1,750 nodes. 69.8\% (698/1,000) of graphs in our MalNet-Tiny splits contain more than 1,500 nodes, including 69.9\%, 71.0\%, and 69.0\% of the training, validation, and test sets, respectively. Moreover, the median training graph contains 2,272 nodes. The benchmark excludes computation of DiGress’s spectral and cycle-count auxiliary features and therefore provides a favorable lower bound on the complete training pipeline’s memory requirements.

\textbf{DeFoG.} DeFoG was not benchmarked on the GH200 because its reference environment targets AMD64/x86-64, whereas our GH200 host uses ARM64/aarch64. We therefore evaluated the DeFoG graph-transformer backbone on an NVIDIA RTX 4090 (23.64 GiB) using its configured architecture (5 layers, hidden node/edge dimensions 256/64, 8 attention heads), batch size 1, and FP32. A complete forward/backward training step succeeded for a 750-node graph, requiring 20.97 GiB peak allocated GPU memory. At 775 nodes, the forward pass succeeded (22.21 GiB peak), but the backward pass produced CUDA OOM. Thus, the observed training-memory limit was between 750 and 775 nodes. Approximately 70\% of MalNet-Tiny graphs exceed 1,500 nodes, making training on the dataset infeasible under this setup.

\textbf{SparseDiff}. Exhibited substantially better scalability than dense diffusion baselines, successfully completing forward and backward passes on a 2,000-node MalNet-Tiny graph using 86.8 GiB peak GPU memory on an NVIDIA GH200 (95 GiB available). However, under the evaluated configuration ($lambda$=0.5, batch size 1, FP32), a 2,999-node graph exceeded available memory during the forward pass. At this setting, the sampled query set grows with the number of possible node pairs: the 2,999-node graph required approximately 2.25M query edges and a 4.50M-edge computational graph.

\textbf{PARD.} We evaluated the PARD PPGN-Transformer backbone using its configured architecture (8 layers, hidden node/edge dimensions 256/64, 8 attention heads) with batch size 1 and FP32 on an NVIDIA GH200 with 95 GiB of available GPU memory. PARD completed inference at 1,500 nodes using 7.85 GiB peak allocated GPU memory, but memory requirements increased rapidly with graph size, reaching 13.83 GiB at 2,000 nodes, 32.63 GiB at 3,000 nodes, 57.97 GiB at 4,000 nodes, and 81.72 GiB at 4,750 nodes. At 5,000 nodes, inference exceeded the available GPU memory, reaching 84.79 GiB peak allocated memory before failing on an additional 23.84 GiB allocation. Thus, the observed inference-memory limit on the 95-GiB GH200 lies between 4,750 and 5,000 nodes. Despite PARD’s autoregressive block decomposition, its dense PPGN-Transformer backbone therefore exhibits severe memory growth with graph size. This benchmark evaluates only the neural backbone in inference mode and consequently provides a favorable lower bound on the memory requirements of the complete training pipeline.

\section{Additional Ordering Analysis}
\label{app:ordering_analysis}

To further characterize the proposed structural ordering, we measure the
average number of distinct node embeddings induced by different structural
ranking schemes. A larger value indicates that the ordering signal separates
more nodes before BFS tie-breaking, reducing the number of structurally tied
nodes during serialization. Table~\ref{tab:ordering_discriminativeness}
compares degree ordering, the PARD structural ordering at depths 1--5, and our
SIR-GN-based ordering with and without node/edge attributes. Across datasets,
our ordering produces substantially more distinct embeddings than degree and
PARD-style ordering, indicating a more discriminative structural ranking signal.

\begin{table}[t]
\centering
\small
\caption{Ordering discriminativeness measured by the average number of distinct
structural embeddings/ranks. Higher values indicate fewer ties in the structural
ranking before BFS tie-breaking. Values for PARD and our orderings are reported
for depths 1--5.}
\label{tab:ordering_discriminativeness}
\setlength{\tabcolsep}{4pt}
\begin{tabular}{llcccccc}
\toprule
Order & Depth & QM9 & ZINC250K & MOSES & ANI-1x & QM7x & Transition1x \\
\midrule
Avg. nodes & -- & 18.07 & 23.09 & 21.60 & 16.57 & 15.02 & 13.62 \\
\midrule
Degree & -- & 3.57 & 3.27 & 3.21 & 3.61 & 3.58 & 3.57 \\
\midrule
PARD ordering & 1 & 4.32 & 5.21 & 5.20 & 4.15 & 3.87 & 3.67 \\
PARD ordering & 2 & 5.59 & 7.57 & 7.28 & 5.61 & 5.07 & 4.73 \\
PARD ordering & 3 & 7.02 & 11.25 & 10.56 & 6.84 & 5.71 & 5.49 \\
PARD ordering & 4 & 7.47 & 12.56 & 11.65 & 7.29 & 5.88 & 5.63 \\
PARD ordering & 5 & 7.52 & 13.07 & 12.01 & 7.39 & 5.89 & 5.64 \\
\midrule
Ours, no attr. & 1 & 9.12 & 12.77 & 15.23 & 10.11 & 8.77 & 9.90 \\
Ours, no attr. & 2 & 12.48 & 18.29 & 19.02 & 12.49 & 10.87 & 9.90 \\
Ours, no attr. & 3 & 13.50 & 20.20 & 19.69 & 13.14 & 11.14 & 10.16 \\
Ours, no attr. & 4 & 13.62 & 20.61 & 19.80 & 13.26 & 11.15 & 10.17 \\
Ours, no attr. & 5 & 13.63 & 20.70 & 19.82 & 13.28 & 11.15 & 10.17 \\
\midrule
Ours, attr. & 1 & 9.45 & 17.53 & 16.47 & 10.34 & 8.91 & 8.15 \\
Ours, attr. & 2 & 12.67 & 21.20 & 19.32 & 12.66 & 11.05 & 10.01 \\
Ours, attr. & 3 & 13.57 & 21.82 & 19.80 & 13.26 & 11.28 & 10.26 \\
Ours, attr. & 4 & 13.68 & 21.93 & 19.88 & 13.37 & 11.29 & 10.27 \\
Ours, attr. & 5 & 13.68 & 21.95 & 19.89 & 13.38 & 11.29 & 10.27 \\
\bottomrule
\end{tabular}
\end{table}

The gap is largest on ZINC250K and MOSES, where degree produces only about
three distinct ranks on average, while our attribute-aware depth-5 ordering
separates roughly 22 and 20 distinct embeddings, respectively. This suggests
that the proposed structural ranking produces a much finer ordering signal than
degree and PARD-style structural neighborhoods. Importantly, this analysis does
not by itself prove improved generation quality; rather, it supports the
mechanism behind the method by showing that the ordering used for serialization
contains substantially more discriminative structural information.

\section{Additional Algorithms}
\label{app:algorithms}

This appendix provides detailed pseudocode for the proposed framework, covering structure-guided ordering, bit-level serialization, edge perturbation, and the two training phases (Algorithms 1–6).

\begin{algorithm}[t]
\caption{Structure-Guided Two-Phase Autoregressive Graph Generation}
\label{alg:pipeline}
\begin{algorithmic}[1]
\Require Training graphs $\mathcal{D}$, graph encoder $\phi$,
         Phase~1 epochs $E_1$, Phase~2 iterations $K$,
         candidates $M$, GMM components $C$, training quantile $q$
\Ensure Trained autoregressive generator $p_\theta$
\State Initialize generator parameters $\theta$

\Statex \textbf{Serialization precomputation}
\ForAll{$G \in \mathcal{D}$}
    \State $\pi_G \gets \textsc{StructureGuidedBFS}(G)$
    \State $s(G) \gets \textsc{BitSerialize}(G,\pi_G)$
\EndFor

\Statex \textbf{Phase 1: exploration-oriented pretraining}
\For{$e = 1,\ldots,E_1$}
    \State Build $\widetilde{\mathcal{D}}_e$ by random edge additions/removals
           on each $G \in \mathcal{D}$
    \State Serialize each $\widetilde{G}\in\widetilde{\mathcal{D}}_e$
           via \textsc{StructureGuidedBFS} and \textsc{BitSerialize}
    \State Update $\theta$ by teacher forcing on
           $\{s(G):G\in\mathcal{D}\}\cup
           \{s(\widetilde{G}):\widetilde{G}\in\widetilde{\mathcal{D}}_e\}$
\EndFor

\Statex \textbf{Phase 2: ReST-style GMM refinement}
\State Fit $\mathrm{GMM}$ with $C$ components on $\{\phi(G):G\in\mathcal{D}\}$
\State $\tau \gets q$-quantile of
       $\{\log p_{\mathrm{GMM}}(\phi(G)):G\in\mathcal{D}\}$
\For{$k = 1,\ldots,K$}
    \State Sample $M$ bit sequences from $p_\theta$ and decode well-formed ones
           into graphs $\{\widehat{G}_j\}$
    \State $\mathcal{C}_k \gets
           \{\widehat{G}_j:
           \log p_{\mathrm{GMM}}(\phi(\widehat{G}_j)) \geq \tau\}$
    \State Serialize each $\widehat{G}\in\mathcal{C}_k$ and continue
           teacher-forced training of $p_\theta$
\EndFor
\State \Return $p_\theta$
\end{algorithmic}
\end{algorithm}

\begin{algorithm}[t]
\caption{Structure-Guided BFS Ordering}
\label{alg:ordering}
\begin{algorithmic}[1]
\Require Graph $G=(V,E)$
\Ensure Node ordering $\pi$ over $V$
\State $\{h_v\}_{v\in V} \gets \textsc{SIR-GN}(G)$
\State Compute structural rank $r(v)$ for each $v\in V$ from $\{h_v\}$
\State $\pi \gets [\,]$; $\mathrm{visited}\gets \emptyset$
\State Sort all nodes by structural rank to obtain component-start order
\While{$|\mathrm{visited}| < |V|$}
    \State Select the highest-ranked unvisited node $v^\ast$
    \State Initialize FIFO queue $Q \gets [v^\ast]$
    \State $\mathrm{inQueue}\gets\{v^\ast\}$
    \While{$Q \neq \emptyset$}
        \State $u \gets \mathrm{front}(Q)$
        \State Append $u$ to $\pi$; $\mathrm{visited}\gets \mathrm{visited}\cup\{u\}$
        \State $\mathcal{N}_u \gets$ neighbors of $u$ sorted by structural rank
        \ForAll{$w\in \mathcal{N}_u$}
            \If{$w\notin \mathrm{visited}$ and $w\notin \mathrm{inQueue}$}
                \State Append $w$ to the back of $Q$
                \State $\mathrm{inQueue}\gets \mathrm{inQueue}\cup\{w\}$
            \EndIf
        \EndFor
        \State Pop $u$ from the front of $Q$
        \State $\mathrm{inQueue}\gets \mathrm{inQueue}\setminus\{u\}$
    \EndWhile
\EndWhile
\State \Return $\pi$
\end{algorithmic}
\end{algorithm}

\begin{algorithm}[t]
\caption{Bit-Level Edge Serialization}
\label{alg:serialize}
\begin{algorithmic}[1]
\Require Graph $G=(V,E)$ with edge attributes, ordering $\pi$
\Ensure Bit sequence $s(G)=(x_1,\ldots,x_T)$
\State Relabel each $v\in V$ by its position in $\pi$
\State $E_{\mathrm{sort}}\gets$ edges sorted lexicographically by $(u,v)$ with $u<v$
\State $b\gets \lceil \log_2 |V| \rceil$ \Comment{bits per node index}
\State $s\gets [\,]$
\ForAll{$(u,v,a)\in E_{\mathrm{sort}}$}
    \State $\beta_u\gets \textsc{Binary}(u,b)$
    \State $\beta_v\gets \textsc{Binary}(v,b)$
    \State $\beta_a\gets \textsc{EncodeAttr}(a)$
    \State Append $\beta_u \Vert \beta_v \Vert \beta_a$ to $s$
\EndFor
\State \Return $s$
\end{algorithmic}
\end{algorithm}
\begin{algorithm}[t]
\caption{Edge-Level Graph Perturbation}
\label{alg:perturb}
\begin{algorithmic}[1]
\Require Graph $G=(V,E)$, drop rate $p_{\mathrm{drop}}$, add rate $p_{\mathrm{add}}$
\Ensure Perturbed graph $\widetilde{G}=(\widetilde{V},\widetilde{E})$
\State $\widetilde{G}\gets \mathrm{copy}(G)$
\State $k_{\mathrm{drop}}\gets
\min(\lfloor p_{\mathrm{drop}}|V|\rfloor, |E|)$
\State $E_{\mathrm{rem}}\gets$ uniformly sample $k_{\mathrm{drop}}$ edges from $E$
\State Remove all edges in $E_{\mathrm{rem}}$ from $\widetilde{G}$
\State $k_{\mathrm{add}}\gets \lfloor p_{\mathrm{add}}|V|\rfloor$
\State $a\gets 0$
\While{$a<k_{\mathrm{add}}$}
    \State Sample distinct nodes $u,v\in\widetilde{V}$ uniformly at random
    \If{$(u,v)\notin\widetilde{E}$}
        \State Add edge $(u,v)$ to $\widetilde{G}$ with the default or empirical edge attribute
        \State $a\gets a+1$
    \EndIf
\EndWhile
\State Remove isolated nodes from $\widetilde{G}$
\State \Return $\widetilde{G}$
\end{algorithmic}
\end{algorithm}

\begin{algorithm}[t]
\caption{Phase 1: Exploration Pretraining with Perturbed Graphs}
\label{alg:phase1}
\begin{algorithmic}[1]
\Require Training graphs $\mathcal{D}$, precomputed sequences
         $\{s(G):G\in\mathcal{D}\}$, epochs $E_1$, perturbation rates
         $p_{\mathrm{drop}},p_{\mathrm{add}}$
\Ensure Pretrained generator $p_\theta$
\State Initialize generator parameters $\theta$
\For{$e=1,\ldots,E_1$}
    \State $\widetilde{\mathcal{D}}_e\gets\emptyset$
    \ForAll{$G\in\mathcal{D}$}
        \State $\widetilde{G}\gets
        \textsc{Perturb}(G,p_{\mathrm{drop}},p_{\mathrm{add}})$
        \State $\widetilde{\mathcal{D}}_e\gets
        \widetilde{\mathcal{D}}_e\cup\{\widetilde{G}\}$
    \EndFor
    \State $\widetilde{\mathcal{S}}_e\gets
    \{\textsc{BitSerialize}(\widetilde{G},
    \textsc{StructureGuidedBFS}(\widetilde{G})):
    \widetilde{G}\in\widetilde{\mathcal{D}}_e\}$
    \State $\mathcal{S}_e\gets
    \{s(G):G\in\mathcal{D}\}\cup\widetilde{\mathcal{S}}_e$
    \ForAll{mini-batches $\mathcal{B}\subset\mathcal{S}_e$}
        \State Update $\theta$ by teacher forcing on $\mathcal{B}$
    \EndFor
\EndFor
\State \Return $p_\theta$
\end{algorithmic}
\end{algorithm}

\begin{algorithm}[t]
\caption{Phase 2: ReST-Style Refinement with GMM Filtering}
\label{alg:phase2}
\begin{algorithmic}[1]
\Require Pretrained generator $p_\theta$, encoder $\phi$, training graphs
         $\mathcal{D}$, iterations $K$, candidates $M$, GMM components $C$,
         training quantile $q$
\Ensure Refined generator $p_\theta$
\State Fit $\mathrm{GMM}$ with $C$ components on $\{\phi(G):G\in\mathcal{D}\}$
\State $\tau\gets q$-quantile of
$\{\log p_{\mathrm{GMM}}(\phi(G)):G\in\mathcal{D}\}$
\For{$k=1,\ldots,K$}
    \State Sample $M$ bit sequences from $p_\theta$ and decode well-formed ones
           into graphs $\{\widehat{G}_j\}$
    \State $\mathcal{C}_k\gets\emptyset$
    \ForAll{$\widehat{G}_j$}
        \State $\ell_j\gets \log p_{\mathrm{GMM}}(\phi(\widehat{G}_j))$
        \If{$\ell_j\ge\tau$}
            \State $\mathcal{C}_k\gets\mathcal{C}_k\cup\{\widehat{G}_j\}$
        \EndIf
    \EndFor
    \State $\mathcal{S}_k\gets
    \{\textsc{BitSerialize}(\widehat{G},
    \textsc{StructureGuidedBFS}(\widehat{G})):
    \widehat{G}\in\mathcal{C}_k\}$
    \ForAll{mini-batches $\mathcal{B}\subset\mathcal{S}_k$}
        \State Continue teacher-forced training of $p_\theta$ on $\mathcal{B}$
    \EndFor
\EndFor
\State \Return $p_\theta$
\end{algorithmic}
\end{algorithm}

\FloatBarrier

\section{Additional Ablations}
\label{app:ablations}

\subsection{Ablation on Phase 1 and the Role of ReST}

To assess whether the components of the proposed training pipeline are
individually necessary, we performed ablation studies on QM9 and
Transition1x. The goal of these experiments is not merely to show that the
full model performs best, but to clarify the functional role of each design
choice across datasets of different difficulty. In particular, the proposed method is built around an
exploration--exploitation decomposition: Phase~1 broadens the support of the
generator through perturbed graphs, while Phase~2 refines this exploratory
behavior through ReST-style filtering. The ablations below test what happens
when one of these ingredients is removed.

\subsubsection{Phase 1 only, with perturbed graphs and without ReST}

\begin{figure}[t]
    \centering
    \begin{subfigure}[b]{0.48\columnwidth}
        \centering
        \includegraphics[width=\linewidth]{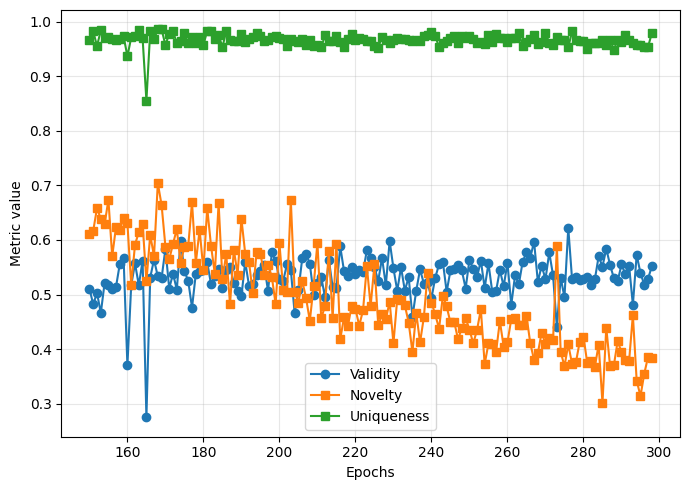}
        \caption{Phase~1 only, with perturbed graphs and without ReST.}
        \label{fig:phase1_rest_ablation_a}
    \end{subfigure}
    \hfill
    \begin{subfigure}[b]{0.48\columnwidth}
        \centering
        \includegraphics[width=\linewidth]{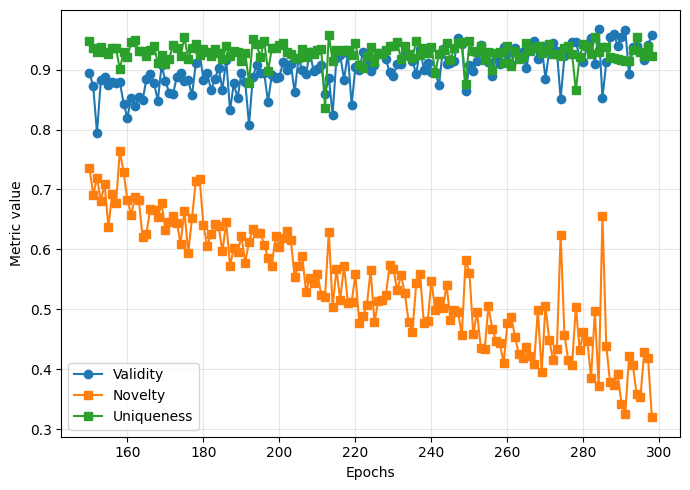}
        \caption{Phase~1 only, without perturbed graphs and without ReST.}
        \label{fig:phase1_rest_ablation_b}
    \end{subfigure}
    \caption{\textbf{Ablation of the two-phase training scheme on Transition1x.} (a) With perturbed graphs
but without ReST, the model explores more broadly but fails to maintain high
validity. (b) Without perturbed graphs and without ReST, the model remains
more conservative but exhibits a steady decline in novelty.}
    \label{fig:phase1_rest_ablation}
\end{figure}

The first ablation considers Phase~1 only, with perturbed graphs included
during training but without the Phase~2 ReST-style refinement. This setting
isolates the exploratory component of the method. As expected, introducing
perturbed graphs broadens the support seen during training and encourages the
generator to move away from a narrow imitation of the empirical distribution.
However, without the exploitative refinement provided by ReST, this broader
support is not filtered back toward plausible regions of graph space. The
result is a generator that explores aggressively but does not sufficiently
specialize, leading to weaker validity and less reliable final samples (Figure~\ref{fig:phase1_rest_ablation}). 
Empirically, this failure is especially clear on Transition1x (Figure~\ref{fig:phase1_rest_ablation_a}), where Validity remains very low and Novelty also deteriorates over time. On QM9, the same ablation maintains relatively high Novelty, but Validity collapses to roughly the 50--60\% range. In both cases, the conclusion is the same: perturbation-based pretraining alone is not enough, and the second phase is necessary to turn exploratory behavior into controlled novelty.

\subsubsection{Phase 1 only, without perturbed graphs and without ReST}
The second ablation removes both ReST and the perturbed-graph augmentation,
leaving only standard Phase~1 training on the original dataset. In this case,
the model no longer benefits from the exploration--exploitation mechanism
introduced by the full method. Instead, it is trained purely to reproduce the
observed training distribution, which leads to a noticeably narrower
generative behavior.

Compared with the previous ablation, this setting maintains substantially
higher Validity, but does so at the cost of reduced exploratory behavior. On Transition1x (Figure 2b), this is reflected in a pronounced decline in Novelty, which drops from roughly 70\% to around 30\% over training. On QM9, the effect is
milder but still visible: Validity and Uniqueness remain near-perfect, while
Novelty stays confined to a relatively narrow range around 88--91\%, far below
the levels reached by the full method. Thus, while this setting remains more
conservative than the perturbed-graph variant, it also fails to generate
sufficiently interesting or diverse samples. This supports the view that the
perturbed graphs are not a superficial augmentation, but the mechanism through
which the model acquires an exploratory bias.

Taken together, these two ablations illustrate the complementary roles of the
two phases. Phase~1 with perturbed graphs encourages exploration but requires
Phase~2 to refine that exploration into valid and distributionally plausible
samples. Phase~1 without perturbed graphs remains too close to pure imitation
and therefore fails to produce the same level of novelty. The full method is
effective precisely because it combines both ingredients. The contrast is sharper on Transition1x than on QM9, suggesting that the benefits of the two-phase exploration--refinement scheme become especially important on structurally harder datasets.

\subsection{Ablation on Node Ordering}

\begin{figure}[t]
    \centering
    \begin{subfigure}[t]{0.48\linewidth}
        \centering
        \includegraphics[width=\linewidth]{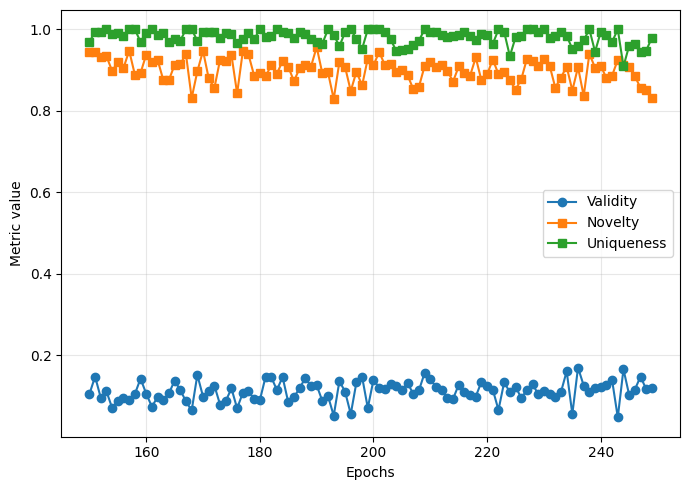}
        \caption{Fully random ordering.}
    \end{subfigure}
    \hfill
    \begin{subfigure}[t]{0.48\linewidth}
        \centering
        \includegraphics[width=\linewidth]{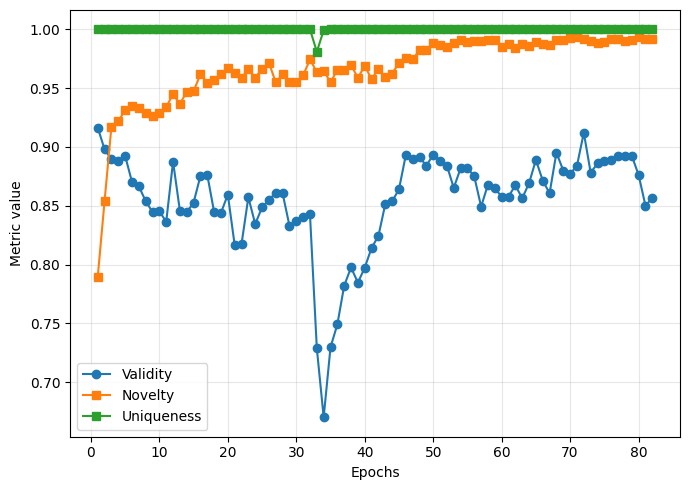}
        \caption{Plain BFS without structural guidance.}
    \end{subfigure}
    \caption{\textbf{Ablation of the node ordering strategy on Transition1x.}
Fully random ordering causes a severe collapse in Validity, while plain BFS
without structural guidance recovers part of the lost structure but remains
weaker than the full structure-guided traversal.}
    \label{fig:ordering}
\end{figure}

To evaluate the importance of the proposed structure-guided serialization, we
compare the full ordering scheme against two weaker alternatives: (i) a fully
random node ordering, and (ii) a plain BFS traversal without structural
guidance. These ablations test whether the gains of the full model genuinely
arise from the ordering strategy, rather than from the autoregressive
architecture alone. For fairness, the final edge list is still sorted
lexicographically after node relabeling, so the ablation changes only the node
ordering and not the final edge-list sorting rule.

\subsubsection{Fully random ordering}

In the first ablation, node identities are assigned through a fully random
permutation, removing both the structural guidance and the BFS constraint. This
produces the weakest possible serialization among the considered variants, since
the autoregressive model no longer receives any meaningful ordering signal.

This ablation leads to the clearest failure mode. On both QM9 and
Transition1x, Validity collapses dramatically, showing that the induced
bit-level sequence becomes too irregular for the decoder to model reliably.
Although the conditional Novelty and Uniqueness scores remain relatively high,
these values must be interpreted with care: Novelty is computed only over the
subset of generated samples that are already valid and unique. Consequently,
high Novelty in this setting does not indicate successful generation, but only
that the small fraction of samples surviving the validity filter are often
unseen. Overall, the random-ordering ablation shows that without a meaningful
serialization, the model fails to produce reliable graphs (Figure~\ref{fig:ordering}).

\subsubsection{Plain BFS without structural guidance}

In the second ablation, we retain the BFS traversal but remove the structural
node ranking used to prioritize candidate nodes during expansion. This
preserves a connectivity-aware traversal bias while isolating the contribution
of the structural information.
The results show that plain BFS is substantially better than a fully random
ordering, confirming that traversal-based locality already provides a strong
inductive bias for sequence generation. On QM9, plain BFS performs comparably
to the full structure-guided ordering, suggesting that on this relatively
regular benchmark the traversal signal alone is already sufficient to induce a
highly learnable serialization. By contrast, on the more challenging
Transition1x dataset, removing structural guidance leads to lower and less
stable Validity, even when Uniqueness and conditional Novelty remain high.
This indicates that structural ranking is not uniformly beneficial across all
datasets, but becomes especially useful when BFS alone is not enough to
organize the serialized sequence effectively.
Taken together, these ablations reveal a clear hierarchy. Fully random ordering
is the least effective and leads to severe validity collapse. Plain BFS
recovers a substantial part of the lost structure and is already competitive on
easier datasets such as QM9. The strongest benefit of the proposed
structure-guided traversal emerges on harder datasets, where BFS alone does not
sufficiently regularize the generation sequence. This suggests that the role of
structural ordering is not simply to replace BFS, but to strengthen it in
regimes where traversal locality alone is insufficient.

\subsection{Ablation on Exploration: Removing Perturbed Graphs}

To isolate the role of the exploratory component of the proposed method, we
consider a variant in which the perturbed graphs of Phase~1 are removed
entirely. In this setting, the model is first pretrained only on the original
training graphs and is then refined through the same Phase~2 ReST-style
procedure. This ablation tests whether the second phase alone is sufficient to
produce strong novelty, or whether the exploratory bias induced by perturbed
graphs is a necessary ingredient of the full pipeline.

The results on Transition1x show that removing perturbed graphs makes the
generator substantially more conservative. During Phase~1, Validity remains
reasonably high, but Novelty steadily declines over training, indicating that
the model increasingly concentrates on a narrow approximation of the empirical
training distribution. In Phase~2, this tendency becomes even more pronounced:
the refinement process further improves plausibility-related metrics, but
Novelty collapses to much lower levels, with only occasional spikes. In other
words, once the exploratory component is removed, the overall pipeline becomes
predominantly exploitative.

This ablation shows that Phase~2 cannot by itself create meaningful novelty; it
can only refine what Phase~1 makes available. Perturbed graphs are therefore
not a secondary augmentation trick, but the mechanism that gives the generator
access to a broader and more exploratory set of graph configurations. Without them, the refinement stage still specializes the model, but largely at
the expense of the novelty that the full method is designed to preserve.

\end{document}